# Recurrent Neural Network Based Modeling of Gene Regulatory Network Using Bat Algorithm

## Sudip Mandal[1*], Goutam Saha[2] and Rajat K. Pal[3]

[1]*Department of Electronics and Communication Engineering, Global Institute of Management and Technology, Krishna Nagar, India.*
[2]*Department of Information Technology, North-Eastern Hill University, Shillong, India.*
[3]*Department of Computer Science and Engineering, University of Calcutta, Kolkata, India.*

*Authors' contributions*

This work was carried out in collaboration between all authors. Author SM designed the study, performed all the simulations and wrote the first draft of the manuscript. Authors GS and RKP gave initial concept of the problem, analyzed the results and revised the manuscript. All authors read and approved the final manuscript.

*Original Research Article*

## Abstract

Correct inference of genetic regulations inside a cell is one of the greatest challenges in post genomic era for the biologist and researchers. Several intelligent techniques and models were already proposed to identify the regulatory relations among genes from the biological database like time series microarray data. Recurrent Neural Network (RNN) is one of the most popular and simple approach to model the dynamics as well as to infer correct dependencies among genes. In this paper, Bat Algorithm (BA) is applied to optimize the model parameters of RNN model of Gene Regulatory Network (GRN). Initially the proposed method is tested against small artificial network without any noise and the efficiency is observed in term of number of iteration, number of population and BA optimization parameters. The model is also validated in presence of different level of random noise for the small artificial network and that proved its ability to infer the correct inferences in presence of noise like real world dataset. In the next phase of this research, BA based RNN is applied to real world benchmark time series microarray dataset of *E. coli*. The results prove that it can able to identify the maximum number of true positive regulation but also include some false positive regulations. Therefore, BA is very suitable for identifying biological plausible GRN with the help RNN model.

_________________________________________

*\*Corresponding author: E-mail: sudip.mandal007@gmail.com;*

# 1 Introduction

Multiple genetic changes which lead to perturbations in specific metabolic pathways that control cell growth are the key reasons for Cancer. In the human body, these alterations often take place owing to mutations in cancer suppressor genes, which in turn leads to uncontrolled cell proliferation, survival and genomic instability. A gene regulatory network (GRN) represents the interactional interrelationships among a group of genes in a cell. The group of genes acts in a synergistic manner to carry out certain functions within the cell. A GRN is usually represented by a graph in which nodes represent genes and regulatory interactions between the genes are represented by directed edges (from the regulator to the regulated gene). The nature of the interaction is of two types, namely activation and repression. Thus, the study of GRNs appears to be very essential in order to discover the genetic causes of a particular disease and for the subsequent design of new and effective methods of treatment to control the disease causing genes with their mutual interactions [1].

DNA microarrays [2,3] are widely used now-a-days for the purpose of investigation of the mechanism that is responsible for gene regulation. Microarrays contain the expression levels of thousands of genes those represent the cause and effect relationship of the species under investigation. This time series microarray database contains underlying information regarding the behavior of genes in terms of changes in their expression in response to cancer causing mutations or any form of treatment at different time sample.

Many types of linear or non-linear mathematical models have been already proposed to infer gene regulatory networks and dynamics from the time series microarray data i.e. a reverse engineering problem. Boolean networks [4,5] examine binary state transition matrices to search patterns in gene expression depending on a binary state function. A Dynamic Bayesian network [6,7] makes conditional probabilistic transitions between network states that merge the features of Hidden Markov model to include the feedback. S-system [8-13] is also a popular model of Biochemical System Theory, represents a GRN as a set of differential equation with power law function. Neural Network [14,15] along with GA was also proposed to infer GRN successfully.

However, in this work, we have used Recurrent Neural Network (RNN) [16,17] which is closed loop Neural Network with a delayed feedback variable suitable to model genetic system dynamics from temporal data. Generally, RNN along with an optimization method is used to infer the GRN where objective function of optimization is the training error. Chiang et al. [18] proposed the hybridization of Genetic Algorithm (GA) and RNN for finding feed-forward regulated genes when some transcription factors were given to construct cancer-related regulatory modules in human cancer microarray data. Xu et al. [19] used Particle Swarm Optimization (PSO) to predict of dynamics and network structure of small artificial network and SOS network of *E. coli.,* but it suffered from its accuracy of the model. Palafox et al. [20] have implemented K-means Population Based Incremental Learning (KPBIL) to optimize the parameters of the RNN and the model is tested against small real and artificial network. The results found to be reliable but it also included few unnecessary regulations in GRN. Noman et al. [21] proposed Decoupled Recurrent Neural Network which was trained Differential Evolution (DE) technique and introduced a penalty term or L1 regularizer in the objective function to balance between accuracy and network structure. The result had shown very good accuracy in finding all true regulation and dynamics for both small and large network. The main disadvantage of this process was inclusion of large number of false regulation. Rakshit et al. [22] have proposed weight matrix based Recurrent Fuzzy Neural Model using Invasive Weed and Artificial Bee Colony (ABC) Optimization technique along with a new penalty function. But results were very poor for both small artificial and real GRN. Kentzoglanakis et al. [23] hybridized PSO and Ant Colony Optimization (ACO) for reverse engineering problem of GRN where PSO was used to train the RNN parameters and ACO was introduced to find the biological plausible network structure. It was tested against small artificial, synthetic and real network with very good accuracy. But the process was very time consuming due to parallel implantation of two separate optimization technique. S. Mandal et al. [24] use a hybrid Cuckoo-search and Flower Pollination algorithm to deal with large scale GRN based on RNN. But it do performed poor for real life network. A. Khan et al. [25] used a swarm intelligence technique to infer small and medium scale GRN along with RNN.

Most of these proposed methods, however, yet to accomplish an accurate inference of small scale real life genetic network. However, few of them were able to found all true regulations but they also detected some false regulations. Moreover, No Free Lunch (NFL) theorem [26] logically states that there is no single metaheuristic which is best suited for solving all kind of optimization problems. Therefore, inference of small genetic network using other optimization techniques is still an open problem to the researchers.

The Bat Algorithm (BA) is very efficient optimization technique which was first proposed by X.-S Yang [27,28]. BA was based on the use of echolocation of bats during their foraging. It has been already reported [27] that BA has superior performances over other metaheuristics like GA and PSO. However, to the authors' best knowledge, Bat Algorithm (BA) was successfully implemented in others field of engineering [29,30] has yet to be incorporated for parameters optimization of RNN. As BA can be successfully implemented for continuous parameter optimization and multi-objective optimization [31,32], it may be suitable for parameter optimization of RNN based model of GRN.

In this paper, BA is introduced for reconstruction of GRN using decoupled and regularized RNN model. The rest of the paper is organized as follows. The preliminary of RNN and BA are discussed in next section. The details of fitness function of BA for decoupled RNN along with penalty or pruning term and learning process for finding accurate structure of GRN are discussed in Section III. After that, the effectiveness of the proposed BA based RNN model is tested against artificial GRN (with and without presence of noise). Results are also compared with others existing techniques. Then a rigorous study was carried out on parametric sensitivity due to variation BA parameters to create best model for GRN. Next, decoupled RNN model with a known regularizer technique is proposed for reconstruction of real life bench mark SOS network for *E. coli*. Conclusion is given in section V followed by references.

## 2 Theoretical Background

Before elaborating the methodology of optimizing the RNN for GRN using Bat Algorithm, let's revise the basic concepts of RNN and BA.

### 2.1 Preliminary of RNN model for GRN

As the inputs of a classical Artificial Neural Network are supplied only from the training dataset, NNs are not suitable to model the dynamics of a system. However, RNN model is a closed loop NN with a delay variable between the outputs of each neuron in the output layer of the RNN to each of the neurons in the input layer that is appropriate to model temporal data. In canonical RNN model [33], the gene's regulations are expressed with the following tightly coupled architecture [16,18-23] where it is assumed that each of the total *N* neurons in the output unit $e_i(t + \Delta t)$ is a gene expression value of next time instant, and the neurons in the input units $e_i(t)$ are the gene expression of present state for same genes, thus they interacts with each and everyone.

$$e_i(t + \Delta t) = \frac{\Delta t}{\tau_i} f\left(\sum_{j=1}^{N} w_{i,j} e_j(t) + \beta_i\right) + \left(1 - \frac{\Delta t}{\tau_i}\right) e_i(t) \qquad \text{where } i = 1, 2, \dots, N \qquad (1)$$

where, $f()$ is a nonlinear function that acts as a classification function. Usually the sigmoid function is used for it where $f(z) = 1/1 + e^{-z}$. $w_{i,j}$ represents the type and strength of the regulatory interaction from *j*-th gene towards *i*-th gene. The positive (negative) value of $w_{i,j}$ represents activation (repression) control of gene-*j* on gene-*i*. $w_{i,j} = 0$ means gene-*j* has no regulatory control on gene-*i*. $\beta_i$ represents the basal expression level and $\tau_i$ denotes the decay rate parameter of the *i*-th gene. $\Delta t$ is incremental time instance, in this work it is set always as 1. So, in the discrete form the RNN model for GRN can be described by the following set of *N(N+2)* unknown parameters $\Omega = \{w_{i,j}, \beta_i, \tau_i\}$ where *i, j* = 1, 2, ···,*N*.

## 2.2 Preliminary of Bat Algorithm for training of RNN

Bat Algorithm (BA), initially proposed by Yang [27], is inspired by echolocation behaviour of bats [34]. Echolocation is one kind of sonar which is used by bats to forage prey and also to avoid obstacles in their path. Bats transmit very loud and high frequency sound continuously and listen for the echo that reflects back from the surrounding objects. Thus a bat can compute direction and distance of the object from the transmitting and receiving wave. Moreover bats can discriminate between a prey and an obstacle easily even in complete darkness [34]. In order to convert these behaviours to Bat Algorithm, some rules are idealized by Yang [28].

- All bats use echolocation to measure distance or direction of objects, and they can also discriminate the difference between food/prey and background obstacles by some magical way [28,34,35].
- Bats fly randomly with velocity $v_i$ at position $x_i$ with a minimum frequency $f_{min}$, varying wavelength $\lambda$ and loudness $A_0$ to search for prey. They can automatically adjust the wavelength (or frequency) of their emitted pulses and adjust the rate of pulse emission $r \in [0, 1]$, depending on the proximity of their target [28,34,35].
- Although the loudness can vary in many ways, it is assumed that the loudness varies from a large (positive) $A_0$ to a minimum constant value $A_{min}$ [28,34,35].

### 2.2.1 Initialization of parameters or solutions

Initial population [28,34] of positions of bats are randomly produced from real-valued vectors with dimension *jb* and number of bats *n*, by taking into account lower and upper boundaries.

$$x_{ib,jb} = x_{min,jb} + rand(0,1) * (x_{max,jb} - x_{min,jb}) \qquad (2)$$

where *ib*=1, 2,…*n* and *jb*=1, 2,…*.d*. $x_{max,jb}$ and $x_{min,jb}$ are lower and upper boundaries for dimension *jb* respectively. *rand* is a function that generate random value within the limit [0,1].

### 2.2.2 Update process of frequency, velocity and position

The frequency factor controls step size of a solution in BA [28,34]. This factor is assigned to random value for each bat (solution) between lower and upper boundaries [$Q_{min}$, $Q_{max}$]. Velocity of a solution is proportional to frequency and new solution depends on its new velocity.

$$Q_{ib} = Q_{min} + (Q_{max} - Q_{min})\beta \qquad (3)$$

$$v_{ib}^{nt} = v_{ib}^{nt-1} + (x_{ib}^{nt} - x^{best})Q_{ib} \qquad (4)$$

$$x_{ib}^{nt} = x_{ib}^{nt-1} + v_{ib}^{nt} \qquad (5)$$

Where *nt* denotes the iteration number, $x^{best}$ denote the current global best solution so far. $\beta \in [0, 1]$ is random generated number to modify the frequency. For local search part of algorithm (exploitation) one solution is selected among the selected best solutions and random walk [28,34] is applied.

$$x_{new} = x_{old} + \varepsilon A^{nt} \qquad (6)$$

Where $A^{nt}$ is average loudness of all bats and $\varepsilon \in [0, 1]$ is random number which represents direction and intensity of random-walk [28,34,35].

### 2.2.3 Loudness and pulse emission rate update process

Loudness and pulse emission rate must be updated as iterations proceed [28,34]. As a bat gets closer to its prey then loudness A usually decreases and pulse emission rate also. Loudness A and pulse emission rate r are updated by the following equations

$$A_{ib}^{nt+1} = \alpha A_{ib}^{nt} \qquad (7)$$

$$r_{ib}^{nt+1} = r_{ib}^{0}[1 - e^{-\gamma * nt}] \tag{8}$$

where $\alpha$ and $\gamma$ are loudness reduction and pulse rate increment constants [28], [34]. $r_{ib}^{0}$ and $A_{ib}^{0}$ are initial pulse rate and initial loudness which are random values between [0,1].

## 3 Materials and Methods

So, the RNN model represents a set of $w_{i,j}$, $\beta_i, \tau_i$ which are called as RNN model parameter. The inference method using the RNN model of genetic network is done by finding the optimum values of RNN parameter with the help of BA such that training error is minimized.

### 3.1 Estimation criteria and decoupled recurrent neural network

All optimization methods use an objective function or a fitness value to measure the goodness of a solution. Most commonly estimation criterion is squared error which is defined as follows

$$f = \sum_{k=1}^{M} \sum_{i=1}^{N} \sum_{t=1}^{T} (e_{cal,k,i,t} - Xe_{exp,k,i,t})^2 \tag{9}$$

where *N* is the number of genes in the problem, *T* is the number of sampling instances of the observed gene expression data and *M* is number of training dataset. $e_{cal,k,i,t}$ is numerically calculated gene expression value of *k*-th dataset at time *t* of *i*-th gene using the set of obtained parameter of RNN model. $e_{exp,k,i,t}$ is the actual gene expression level of *k*-th dataset at *t*-time of *i*-th gene. The *f* denotes total squared error between the calculated and the observed gene expression data gene expression. Therefore, RNN modeling is a non-linear function optimization problem to discover the optimal RNN parameter by minimizing the fitness function or mean square error so that calculated gene expression data fits with the observed gene expression data. Since for *N* genes, *N(N+2)* parameters must be determined to find solution of set of equations (1), thus the RNN model method search for the best RNN parameters values in *N(N+2)* dimensional space. This space is too large dimensional in cases of large-scale genetic networks that consist of many genes. So, to overcome this problem, here the genetic network inference problem is divided or decoupled into several sub-problems corresponds to each gene. Now, the objective function of the sub-problem corresponding to *i*-th gene is to find the values of decoupled RNN model parameters which minimizes error for only *i*-the gene $f_i$ and it is defined as follows

$$f_i = \sum_{k=1}^{M} \sum_{t=1}^{T} (e_{cal,k,i,t} - e_{exp,k,i,t})^2 \tag{10}$$

$$e_i(t + \Delta t) = \frac{\Delta t}{\tau_i} f\left(\sum_{j=1}^{N} w_{i,j} e_j(t) + \beta_i\right) + \left(1 - \frac{\Delta t}{\tau_i}\right) e_i(t) \tag{11}$$

Hence, to solve the differential equation (11), the number of RNN parameters needed to determine is only *(N+2)* parameters for *i*-th gene. Thus, this decoupling method divides a *N(N+2)*-dimensional problem space into *(N+2)*-dimensional sub problem space for each gene. By accumulating the *(N+2)* parameters of all *N* genes, overall structure of RNN can be achieved which in turn denotes the GRN.

### 3.2 Regularization as penalty term for real life network

However, real life genetic network are sparsely connected i.e. very few connectivity exist among genes and the measured data are also very noisy. So, RNN based GRN model may have different optimal solutions with very low error value depending on the different connectivity or structures among those genes of the network and the corresponding values of the kinetic parameters. This is known as over-fitting problem. So, to overcome this over-fitting problem for real life genetic network, the balance between the minimization of error and actual regulatory structure of GRN need to be achieved. So, a regularization term is introduced as penalty term along with error function to avoid over-fitting, to find out the original network by restricting regulator size and also to restrict reach space during optimization.

To generate the sparse solutions, the concept of in-degree or cardinality [21] of genes in error function was already introduced. Cardinality of gene is defined as the allowed number of regulations over the particular genes. In this paper, a penalty term based maximum cardinality *I* is added to fitness function for real life network reconstruction where it is assumed that out of *N* kinetic parameter values of each of the *w*, only *I* non-zero values are allowed within each *w* vectors, thus forcing the other (*N-I*) values to become zero. If any of these (*N-I*) elements achieved a non zero-value during optimization process, the solution will be penalized in the following way [21] for decoupled RNN system

$$f_i = \sum_{k=1}^{M} \sum_{t=1}^{T} (e_{cal,k,i,t} - e_{exp,k,i,t})^2 + c \sum_{j=1}^{N-I}(|W_{i,j}|) \qquad (12)$$

where $W_{i,j}$ is the vector which contains the absolute values of $w_{i,j}$ but sorted in ascending order. *C* is the weight constant that denote magnitude of penalty to balance between over fitting and actual network structure.

### 3.3 Learning process

Learning the optimal values of parameters that best fit with the training data is an optimization problem. So, to optimize the parameters of RNN, BA was introduced by minimize the error or fitness function $f_i$. In Bat Algorithm, velocity and position of a bats is corresponds to a solution (i.e. a set of $w_{i,j}$, $\beta_i$, $\tau_i$) in search space and it moves towards minimum error area gradually during iteration. Now the pseudo code [27], [28], [34] of the proposed method can be given as

---

***Start BA based RNN***
*Define Objective function f(x)= $f_i$ (MSE error for Decoupled RNN along with the penalty term and value of I, nt, n, d, gn) where n is the number of initial solutions or bats, d is the dimension of problem, I is the Cardinality and nt is number of iteration, gn is gene index.*

*for i=1: n*
    *Generate initial population of Bat positions $x_{ib}$ and velocity $v_{ib}$ with the n set of [$w_{ib,jb}$, $\beta_{ib}$, $\tau_{ib}$] (for ib = 1, 2, ..., n and jb= 1,2,...,d) with proper range*
*end for;*

*Initialize pulse frequency $Q_{ib}$, pulse rate $r_{ib}$ and loudness $A_{ib}$ at $x_{ib}$.*
*Find out the current best solution.*

*for i=1:nt*
    *for i1=1:n*
        *Generate new solutions by adjusting frequency, and updating velocities and location/solutions.*
        *if (rand > $r_{ib}$)*
            *Select a solution among the best solutions*
            *Generate a local solution around the selected best solution.*
        *end if*
        *if (rand< $A_{ib}$ and f($x_{ib}$) < f($x^{best}$))*
            *Accept new solutions.*
            *Increase $r_{ib}$, reduce $A_{ib}$*
        *end if*
        *Ranks the bats and find current best $x^{best}$*
    *end for*
*end for*

*Display results.*
***End BA based RNN***

---

# 4 Results and Discussion

In this research work, the performance of the inference algorithm was evaluated over the both artificial and real networks and experimental results were compared with the others existing methods. Now, the performance of the modeling of GRN is measured by two processes.

First most important performance criterion is measured from network structure or architecture point of view where inferred network is compared with the original network structure with respect to edge connectivity. Now, the two parameters, Sensitivity ($S_n$) and Specificity ($S_P$) of the reconstructed network are define as follows [11,20, 21].

$$S_n = \frac{TP}{TP+FN}, \quad S_p = \frac{TN}{TN+FP} \tag{13}$$

where TP (True Positive) denotes the number of correctly predicted regulations, TN (True Negative) represents the number of properly predicted non-regulations, FP (False Positive) denotes the number of incorrectly predicted regulations and FN (False Negative) represents the number of falsely predicted non-regulations by the inference algorithm.

The inferred values of the regulatory parameters are also bit of concern as its magnitude can affect the network connectivity. So, we define another performance measurement parameter as Inferred Parametric Error (*IPE*) which measures the deviation in magnitude of inferred parameters of RNN model from original one.

$$IPE = \prod_{i,j=1}^{N} \left| w_{i,j}^{exp} - w_{i,j}^{cal} \right| \tag{14}$$

Where $w_{i,j}^{exp}$ is the actual value of weight parameter and $w_{i,j}^{cal}$ are the calculated value of the same. Using these two types of performance parameter, we can estimate the efficiency of an inferred algorithm. All the experiments were performed using computer with an Intel© Dual Core processor, 2GB of RAM, Windows 7 platform and Matlab R2009b software tools.

## 4.1 Inference for small artificial network with noise free data

To taste the effectiveness of the Bat Algorithm based RNN model of GRN, a known small artificial regulatory network was chosen which contain four genes with simple regulatory dynamics. Earlier approaches [19,21,22] already used this network to verify their proposed algorithm's efficiency.

#### 4.1.1 Experimental setup

First the proposed methodology applied to noiseless data using the parameters which are shown in the Table 1. If an insufficient amount of time-series data is given as training data, due to the high degree-of-freedom of RNN, several optimal solutions may be exist. However, in this paper an ample amount of time-series data is used to improve the chances of finding the correct interactions.

Specifically, here, 6 sets of noise-free time-series data were used each consist of all 4 genes. The time-series data were generated by solving the set of differential equations (1) and the initial values of these sets were selected randomly. In real life, these time-series data could be obtained by different biological experiments. The number of sampling point was set as 15. For our work, there were total 90 data points for each of the genes. For each gene, 6 parameters need to be identified by using BA. The search space was selected as $w_{i,j} \in$ [-30,30], $\beta_i \in$ [-10,10] and $\tau_i \in$ [0, 20] same as earlier work [21].

Cardinality *I* was set as 4 i.e. the maximum number of all types of possible regulations for a particular gene was 4 and value of *C* was set as 1. For BA, *nt*, maximum number of iteration was 2000 and initial number of solutions was 200. Value of *α* and *γ* was set as 0.1 and 0.1 respectively which were chosen after testing with some possible values that will be discussed later. Boundary of frequency was initialized as [0,1] and step size during random walk was fixed by 0.001. The state of the rand function was reset to get the fixed output for all program execution.

### 4.1.2 Results

Table 2 shows the inferred parameters for the first attempt of experiment. If the values of kinetic orders which have absolute value less than 0.05 were ignored, the BA based RNN model gave quite satisfactory results for noiseless data as it gave almost correct values of all parameters. Moreover, BA can also able to infer the correct signs and positions of regulations and non-regulations accurately. In spite of that, it can be concluded that if sufficient number of large time series data was available, BA can able to detect all correct values and directions of regulations very efficiently with negligible variations in numerical values.

The main advantages of this BA based RNN model were the reduction in data point, minimum *IPR* and faster convergence. It was observed that BA was converged within the 500 iteration for 200 bats to reach the stability or minimum error which is around $1 \times 10^{-7}$ for all genes; while others evolutionary methods needed more number of iterations and dataset. The comparative study among existing methods to infer noise free small RNN based artificial network were shown in Tables 2 and 3.

It can be observed that the proposed RNN-BA method is much superior than RNN-IWO+ABC [22] and RNN-ACO+PSO [23] with respect to selectivity and sensitivity. Moreover from Table 3, it can be seen that RNN-BA model can reconstruct small artificial network with least number of training data sample point. *IPE* of RNN-BA model is also very smaller and better than RNN-IWO+ABC [22] and RNN-ACO+PSO [19] that proves the accuracy of the proposed model. However, only RNN-DE [21] process gave better performance than our RNN-BA model in term of IPR. In overall, RNN-BA is very promising method to model small and noiseless artificial network.

**Table 1. Actual RNN model parameters for small artificial network**

| $w_{i,j}$ | 1  | 2   | 3  | 4   | $\beta_i$ | $\tau_i$ |
|-----------|----|-----|----|-----|-----------|----------|
| 1         | 20 | -20 | 0  | 0   | 0         | 10       |
| 2         | 15 | -10 | 0  | 0   | -5        | 5        |
| 3         | 0  | -8  | 12 | 0   | 0         | 5        |
| 4         | 0  | 0   | 8  | -12 | 0         | 5        |

**Table 2. Inferred RNN Parameters for Artificial Network using Noiseless Data**

| $w_{i,j}$ | 1     | 2      | 3     | 4      | $\beta_i$ | $\tau_i$ |
|-----------|-------|--------|-------|--------|-----------|----------|
| 1         | 19.97 | -19.96 | -0.01 | 0.03   | -0.01     | 10.00    |
| 2         | 14.99 | -9.99  | 0.00  | 0.00   | -5.00     | 5.00     |
| 3         | -0.01 | -7.99  | 12.00 | 0.01   | 0.00      | 5.00     |
| 4         | 0.00  | 0.00   | 8.00  | -11.99 | 0.00      | 5.00     |

**Table 3. Comparative study-I among Existing Methods for Noise Free Artificial Network**

| Process          | TP | FP | TN | FN | $S_n$ | $S_p$ |
|------------------|----|----|----|----|-------|-------|
| RNN-BA           | 8  | 0  | 8  | 0  | 1     | 1     |
| RNN-PSO [19]     | 8  | 0  | 8  | 0  | 1     | 1     |
| RNN-DE [21]      | 8  | 0  | 8  | 0  | 1     | 1     |
| RNN-IWO+ABC [22] | 7  | 3  | 5  | 1  | 0.88  | 0.63  |
| RNN-ACO+PSO [23] | 5  | 1  | 7  | 3  | 0.63  | 0.88  |

## 4.2 Inference for artificial system with noisy data

In the next phase of the experiment, the proposed method was tested against noisy artificial data to test whether the model was noise immune or not, as in real life data there are lots of measurement errors or noise. Here, the added noise was random in nature and different percentage of randomness was added to observe the performance with the increase of effect of noise in the data. The noise was added in the following way

$$Noisy(d) = d(t) * [(1 - \frac{ns}{100}) + 2 * rand * \frac{ns}{100}] \quad (15)$$

Where *d(t)* is initial noiseless data, *ns* is percentage of random noise and *rand* is a function that generate random number between [0,1]. As *ns* increases, the *Noisy(d)* loses the originality from the actual data. In this paper, the results for *ns* = 5%, 10% random noises were observed. The optimization search parameters and the regularization parameters setting were same as in the previous noise-less case.

**Table 4. Comparative study-II among existing methods for noise free artificial network**

| Process | No of training dataset | No of time sample | Total data point used | IPE |
|---|---|---|---|---|
| RNN-BA | 6 | 15 | 90 | 0.14 |
| RNN-PSO [19] | 3 | 50 | 150 | 608.5 |
| RNN-DE [21] | 10 | 50 | 500 | $\cong 0$ |
| RNN-IWO+ABC [22] | 4 | 150 | 600 | 45.42 |
| RNN-ACO+PSO [23] | 1 | 300 | 300 | -- |

#### 4.2.1 Results

Table 5 shows the inferred parameters value for noisy system of 4 genes artificial network. It was observed that adding 5% noise does not affect the accuracy of the structure; the value of actual nonzero regulatory parameters remained almost same as earlier noise-free case. But, due to noise we got two prominent false positive regulations whose values were greater than 1, else the magnitudes of others zero-valued non regulatory parameters were deviated with acceptable range. For 10% case, the inference was still successful to detect the all actual regulations and parameters which were still in acceptable range, but 3 FPs were added to the network. It is interesting to observe that the number of FPs increased with noise but still number of FN was zero. So, these results proved the robustness against noise of RNN modeling of GRN using BA.

**Table 5. Inferred RNN model parameters for small artificial network using noisy data**

| Noise level | $w_{i,j}$ | 1 | 2 | 3 | 4 | $\beta_i$ | $\tau_i$ | TP | FP | TN | FN |
|---|---|---|---|---|---|---|---|---|---|---|---|
| 5% | 1 | 19.06 | -17.74 | 1.27 | 1.06 | -1.93 | 9.55 | 8 | 2 | 6 | 0 |
|  | 2 | 14.76 | -9.59 | 0.13 | -0.54 | -4.94 | 4.80 |  |  |  |  |
|  | 3 | 0.82 | -7.47 | 11.40 | 0.37 | -0.89 | 4.65 |  |  |  |  |
|  | 4 | 0.12 | -0.16 | 8.06 | -11.80 | -0.12 | 4.89 |  |  |  |  |
| 10% | 1 | 18.91 | -16.75 | 1.90 | 1.67 | -3.04 | 9.27 | 8 | 3 | 5 | 0 |
|  | 2 | 14.44 | -9.10 | 0.20 | -0.90 | -4.88 | 4.59 |  |  |  |  |
|  | 3 | 1.17 | -8.17 | 12.21 | 0.39 | -1.07 | 4.84 |  |  |  |  |
|  | 4 | 0.23 | -0.30 | 8.02 | -11.41 | -0.27 | 4.77 |  |  |  |  |

### 4.3 Model selection or parametric sensitivity

Next, parametric sensitivity of BA for 4 genes artificial network were observed by varying different parameters of it so that suitable model can be selected.

#### 4.3.1 Population

First, we tested the sensitivity of the population size where we observed the variation of Inferred Parametric Error (*IPE*) or deviation in parameters values with the different iteration number. From Fig. 1, it can be observed that *IPE* was rather insensitive to iteration number but it decreased with initial population size. It is obvious that if number of population or different initial solutions is increased, the optimization technique will perform better.

But it can be also observed that *IPE* was saturated after some value of population (here 200) which denotes minimum population required for the process. It also signifies that though we increase iteration number, *IPE* can't go beyond a boundary level which is the limitation of BA. So, we have chosen minimum 200 populations for all other experiment.

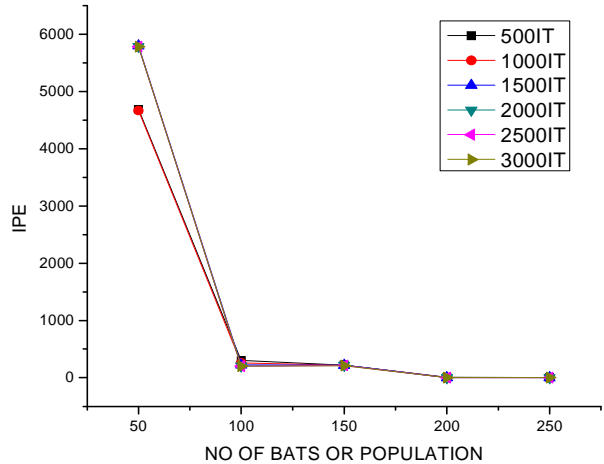

**Fig. 1. Variation of IPE vs. no. of population**

### 4.3.2 Iteration range

Now, we observed the change in fitness value with iteration number of the algorithm which is plotted in Fig. 2. It was observed, fitness value was initially decreasing with the iteration number and then saturated after some point. For all cases of different population, after 2000 iteration, the fitness value got saturated. But, when we used 50 populations, it was noticed that fitness value never went close to zero. So, it can be concluded that both number of iteration and initial population should be large enough for correct inference of RNN model.

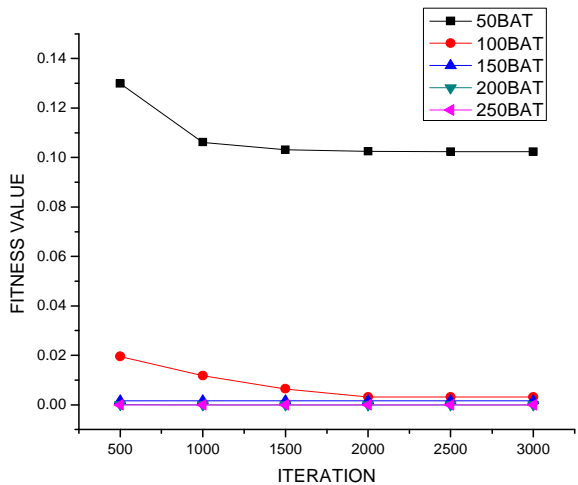

**Fig. 2. Variation of fitness value vs. no. of iteration**

### 4.3.3 Number of time series data

In next part, we have studied the effect of number of time series data to the structure of the network. Figs. 3 and 4 have shown the variation of FP and TN with the increase of time series data for number of iteration 500. When we used a single time series data, the actual network structure could not be obtained where there was presence of extra regulations i.e. false positive. But as the number of time series data was increased there was sharp fall and increase in the number of FP and TN respectively. It was interesting to observed that in all cases of inference, BA always detected all actual regulations of the network i.e. TP is always constant (here it is 8) and consequently FN was always zero. So, sufficient number of time series with different initial sample point is necessary for correct inference of BA.

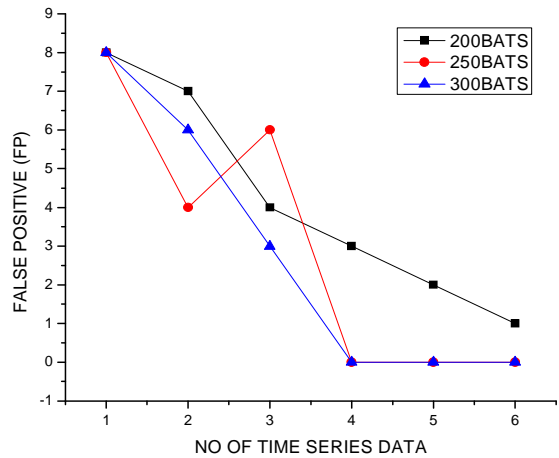

**Fig. 3. Variation of FP vs. no. of time series**

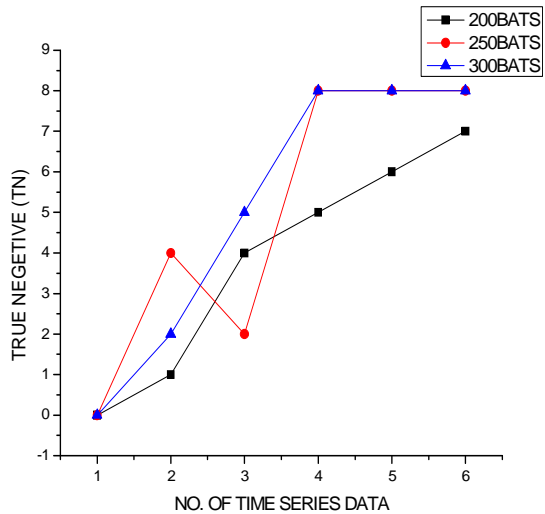

**Fig. 4. Variation of TN vs. no. of time series**

### 4.3.4 Value of α and γ

For BA, *α* and *γ* are two important parameters that make balance between local and global search. Depending on the values of these parameters, convergence rate will be faster or slower. So, the selection of the value *α* and *γ* are very crucial for better performance of BA. Moreover, we observed in our experiment, that if the fitness value is less than equal to $1 \times 10^{-7}$, the BA give very good results in term of RNN model parameters values. So, we conducted our experiment in such way that program will be stopped if the fitness value become equal less than $1 \times 10^{-7}$ and we repeated this experiment with different value of *α* and *γ*. Following table shows the required iteration number with some fixed variation in *α* and *γ*. It is found that for *α*=0.1 and *γ*=0.1 gave best performance. Therefore, we select *α* = 0.1 and *γ*=0.1 for all other experiments.

**Table 6. Minimum iteration required vs alpha and gamma**

| Minimum iteration required | | ALPHA | | | | |
|---|---|---|---|---|---|---|
| | | 0.1 | 0.2 | 0.4 | 0.6 | 0.9 |
| GAMMA | 0.1 | 432 | 484 | 1124 | 877 | 963 |
| | 0.2 | 449 | 474 | 1150 | 929 | 994 |
| | 0.4 | 483 | 571 | 773 | 1378 | 1066 |
| | 0.6 | 519 | 637 | 2333 | 1547 | 1156 |
| | 0.9 | 629 | 658 | 1000 | 1543 | 1326 |

## 4.4 Real time SOS network

The SOS network for *E. coli*. [36] was first introduced by Uri Alon group [37] which is a benchmark in GRN problem to find out the effectiveness of the inference algorithm on real time dataset and network. In SOS network, 8 genes were considered (uvrD, lexA, umuD, recA, uvrA, uvrY, ruvA and polB). During their experiments, DNA of *E. coli*. was irradiate with the UV light, which affected some gene, after that, the network would repair itself by suppressing others genes expression value (Fig. 5). They performed four experiments for different UV light intensities. Each experiment consists of 50 time steps spaced by 6 minutes for each of the eight genes. Since two of them (uvrY and ruvA) have trivial activity in comparison with the rest of the genes in the network, many researchers chosen rest of 6 genes only. But, in this work, all 8 genes were, considered for better transparency and dataset for experimental condition 1 [38] was chosen only for the validation of the proposed model as others dataset were processed under different experimental condition.

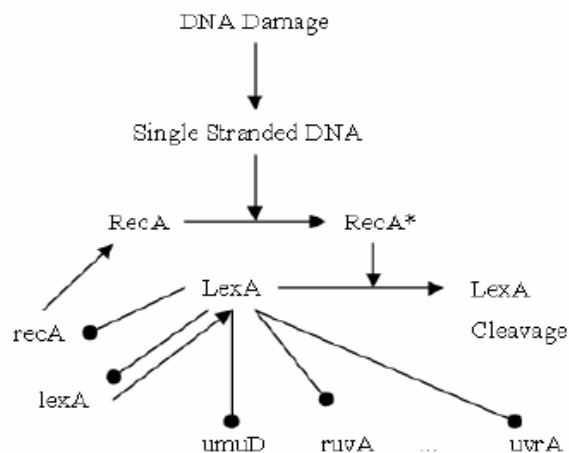

**Fig. 5. Graphical representation of the actual SOS network for *E. coli*. [39]**

For each gene, 10 parameters are needed to be identified by using BA. The search space was selected as $w_{i,j} \in [-10,10]$, $\beta_i \in [-10,10]$ and $\tau_i \in [0, 10]$ same as earlier work [21]. Cardinality *I* was set as 2 i.e. the maximum number of all types of possible regulations for a particular gene was 2 and value of C was set as 1. For BA, *nt*, maximum number of iteration was 2000 and initial number of solutions was 500. Value of *α* and *γ* was set as 0.1 and 0.1 respectively. The value of *dt* was 1. The first time point data was not considered since few data points were zero at that time instance. Moreover, the dataset was normalized within the interval [0,1] before use.

However, due to the noise in data of SOS network for *E. coli*. and the random initialization of metaheuristic, different results (i.e. values of RNN parameters) were obtained for each run. So, the program was executed 15 times and statistical analysis was performed for the inference GRN. Initially, individual standard deviation ($\sigma$) and mean ($\mu$) for each weight parameters were calculated from the set of results corresponds to 15 run. Then, probability of occurrence for a regulation is calculated from conventional normal distribution function by setting $x$ = overall mean of all weights. Next, the results were filtered with a threshold of $p_a$= 0.5 i.e. if the probability of occurrence was greater than or equal to 0.5 then it will be considered as existing regulation of SOS Network else it will be discarded form the network. Table 7 shows the ultimate inferred network in term of Connectivity Matrix. An element of this matrix, $CM_{i,j}$, denotes the regulatory relationship between the *i*-th gene and the *j*-th gene. $CM_{i,j}$ can have values of +1, 0, and -1 depending on the value and sign of the weights of the RNN model. +1 denotes activation, 0 denotes no regulation and -1 denotes suppression.

**Table 7. Inferred SOS Network Using RNN-BAT**

| Gene-id | uvrD | lexA | umuDC | recA | uvrA | uvrY | ruvA | polB |
|---------|------|------|-------|------|------|------|------|------|
| uvrD    | 1    | -1   | 0     | 0    | 0    | 0    | 1    | 1    |
| lexA    | 0    | -1   | 0     | 0    | 0    | 1    | 0    | 0    |
| umuDC   | 1    | -1   | 0     | 0    | 0    | 0    | 1    | 0    |
| recA    | 0    | 0    | 0     | 0    | 1    | 1    | 0    | 0    |
| uvrA    | 0    | -1   | 1     | 1    | 0    | 1    | 0    | 0    |
| uvrY    | 1    | -1   | 0     | 0    | 0    | 0    | 1    | 0    |
| ruvA    | 0    | -1   | 1     | 0    | 1    | 0    | 0    | 1    |
| polB    | 0    | -1   | 0     | 0    | 0    | 0    | 0    | 1    |

**Table 8. Comparative study among existing methods for inference of SOS network**

| Process          | TP | FP | TN | FN |
|------------------|----|----|----|----|
| RNN-BA           | 7  | 16 | 39 | 2  |
| RNN-PSO [19]     | 7  | 2  | 53 | 2  |
| RNN-DE [21]      | 6  | 11 | 44 | 3  |
| RNN-IWO+ABC [22] | 4  | 47 | 6  | 5  |
| RNN-ACO+PSO [23] | 3  | 10 | 44 | 6  |

Table 7 shows the inferred matrix for the *E. coli* SOS Network using RNN and BAT methodology. From that, we find that our proposed method can found 7 true positive regulations among total 9 regulations. Comparative study between existing techniques is given in Table 8 in term of TP and FP. It can be observed that the proposed method is superior than RNN-DE [21], RNN-IWO+ABC [22] and RNN-ACO+PSO [23] with respect to detection of TP. But it also includes lots of unknown and unexpected regulations which may be due to side effect of noise. So, this proposed model can also able to infer real life network with good accuracy.

# 5 Conclusion

Recurrent neural network (RNN) model is a very popular candidate for inferring the GRN from microarray gene expression data in terms of biological plausibility and computational efficiency. In this work, we have implemented the decoupled RNN model where the regulatory parameters of each gene are calculated independently in separate search instances. We incorporate a new metaheuristic, called Bat Algorithm (BA), for inferring the regulators of RNN based GRN along with a regularizer or penalty term in objective function will help to avoid the FPs' creation, while considering a sparse set of network parameters.

To prove the efficiency of this inference algorithm, it was applied to a benchmark problem of artificial network of 4 genes with and without noise. With the use of fewer data points, BA based RNN can infer the network with very high accuracy. But in presence of noise, the number of FPs increased significantly with increment of noise but still it can identify all TPs with good accuracy. It was also found that its noise robustness performance was better than few others existing methods. Next, this inference algorithm was applied to infer real life problem of SOS network and this method can detect the 6 correct regulations but it also includes many unnecessary regulation in the network.

So, this method can infer the GRN with great accuracy. However, the model inferred some false regulations due to noisiness of real life microarray dataset. Different regularization technique may be employed to improve further accuracy and higher speed.

## Competing Interests

Authors have declared that no competing interests exist.